\pdfoutput=1

\documentclass[11pt]{article}

\usepackage{naacl2021}

\usepackage{times}
\usepackage{latexsym}
\usepackage{colortbl}
\usepackage[T1]{fontenc}

\usepackage[utf8]{inputenc}

\usepackage{microtype}

\usepackage{algorithm}
\usepackage{algorithmic}
\usepackage{graphics}
\usepackage{graphicx}
\usepackage{booktabs}
\usepackage{multirow}
\usepackage{makecell}
\usepackage{caption}
\usepackage{hyperref}
\usepackage{url}

\usepackage{amsmath,amsfonts,bm}

\def\eqref#1{equation~\ref{#1}}

\def\1{\bm{1}}

\DeclareMathAlphabet{\mathsfit}{\encodingdefault}{\sfdefault}{m}{sl}
\SetMathAlphabet{\mathsfit}{bold}{\encodingdefault}{\sfdefault}{bx}{n}

\newcommand{\B}[1] {\boldsymbol{#1}}

\def\bm{{\B{m}}}

\def\bs{{\B{s}}}
\def\bt{{\B{t}}}

\def\bx{{\B{x}}}
\def\by{{\B{y}}}

\usepackage{pifont}
\newcommand{\cmark}{\ding{51}}%
\newcommand{\xmark}{\ding{55}}%

\usepackage{color}

\makeatletter
\newcommand{\thickhline}{%
    \noalign {\ifnum 0=`}\fi \hrule height 1pt
    \futurelet \reserved@a \@xhline
}
\makeatother

\title{SPLAT: Speech-Language Joint Pre-Training for Spoken\\Language Understanding}

\author{Yu-An Chung$^1$\thanks{$\;\;$Equal contribution. The work was done when Yu-An Chung was interning in Microsoft.}$\;$, Chenguang Zhu$^{2*}$, Michael Zeng$^2$\\
\textsuperscript{\rm 1}MIT Computer Science and Artificial Intelligence Laboratory\\
\textsuperscript{\rm 2}Microsoft Cognitive Services Group\\
\texttt{andyyuan@mit.edu,\{chezhu,nzeng\}@microsoft.com}}

\begin{document}
\maketitle
\begin{abstract}
Spoken language understanding~(SLU) requires a model to analyze input acoustic signal to understand its linguistic content and make predictions.
To boost the models' performance, various pre-training methods have been proposed to learn rich representations from large-scale unannotated speech and text.
However, the inherent disparities between the two modalities necessitate a mutual analysis.
In this paper, we propose a novel semi-supervised learning framework, SPLAT, to jointly pre-train the speech and language modules.
Besides conducting a self-supervised masked language modeling task on the two individual modules using unpaired speech and text, SPLAT aligns representations from the two modules in a shared latent space using a small amount of paired speech and text.
Thus, during fine-tuning, the speech module alone can produce representations carrying both acoustic information and contextual semantic knowledge of an input acoustic signal.
Experimental results verify the effectiveness of our approach on various SLU tasks.
For example, SPLAT improves the previous state-of-the-art performance on the Spoken SQuAD dataset by more than~10\%.
\end{abstract}

\section{Introduction}
\label{sec:intro}
Spoken language understanding~(SLU) tackles the problem of comprehending audio signals and making predictions related to the content.
SLU has been widely employed in various areas such as intent understanding~\citep{tur2011spoken,bhargava2013easy,ravuri2015recurrent,lugosch2019speech}, question answering~\citep{lee2018odsqa,chuang2020speechbert}, and sentiment analysis~\citep{zadeh2018multimodal}.
Early approaches leverage a two-step pipeline: use automatic speech recognition~(ASR) to transcribe input audio into text, and then employ language understanding models to produce results.
However, such cascaded system has several drawbacks.
First, the transcription produced by the ASR module often contains errors, which adversely affects the language understanding module's prediction accuracy.
Second, even if the transcription is perfect, the rich prosodic information of speech~(e.g., tempo, pitch, and intonation) is inevitably lost after ASR.
In comparison, humans often leverage these information to better understand and disambiguate the content.
Therefore, there has been a rising trend of end-to-end approaches to retain information from audio signals to carry out the understanding task~\citep{serdyuk2018towards,chen2018spoken,haghani2018audio}.

While end-to-end SLU methods are effective, they often suffer from a shortage of labeled training data, especially when the target task is in a novel domain.
One solution is to leverage self-supervised training as is done in pre-trained language models.
Examples like BERT~\citep{devlin2019bert}, GPT~\citep{radford2018improving}, and RoBERTa~\citep{liu2019roberta} are first pre-trained on large-scale unannotated text in a self-supervised fashion to learn rich textual representations before being fine-tuned on downstream tasks with a modest amount of labeled data.
Borrowing this idea, several pre-training methods have been proposed for speech, e.g., wav2vec~\citep{schneider2019wav2vec,baevski2020vq}, contrastive predictive coding~\citep{oord2018representation,riviere2020unsupervised}, autoregressive predictive coding~\citep{chung2019unsupervised,chung2020vector,chung2020improved}, and DeCoAR~\citep{ling2020deep,ling2020decoar}, to capture contextual representations from unlabeled speech data.
Nevertheless, these methods leverage only acoustic data and mainly focus on modeling the acoustic information during pre-training.
As a result, the produced representations may not be optimal for language understanding tasks.

\begin{figure*}[htbp]
  \centering
  \includegraphics[width=\textwidth]{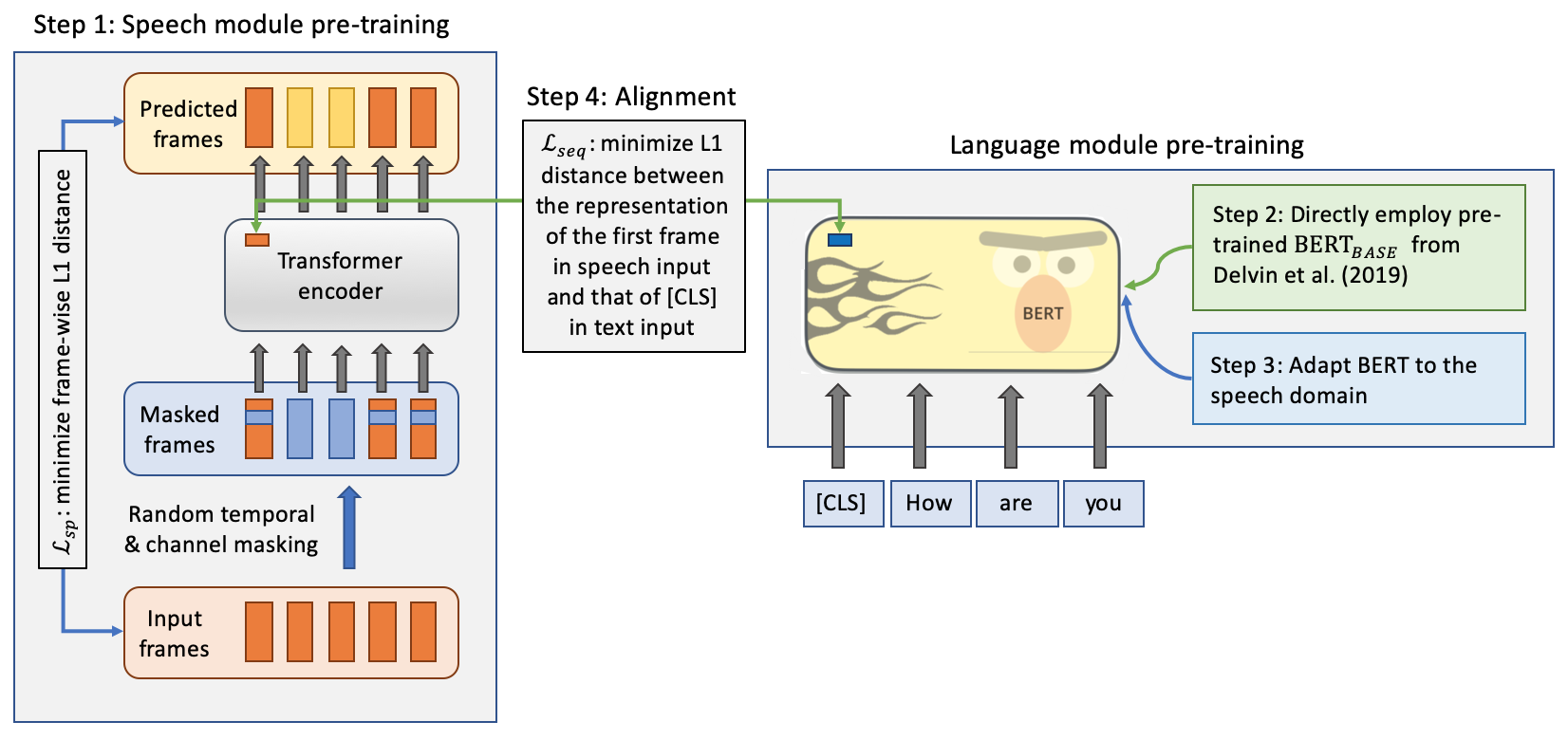}
  \caption{Overview of SPLAT. First, the speech and language modules are separately pre-trained using speech and text data via masked language modeling~(MLM). In practice, we directly employ the BERT$_{\mathrm{BASE}}$ model released by~\citet{devlin2019bert} to be the language module. Then, by leveraging a small amount of paired speech and text data, either a sequence-level alignment loss~$\mathcal{L}_{seq}$ or a token-level alignment loss~$\mathcal{L}_{tok}$ is applied to align the representations from both modules in a shared latent space~(only $\mathcal{L}_{seq}$ is shown here). During alignment, the language module is kept frozen and only the speech module is updated. Before aligning the two modules, there is an optional step to update the BERT$_{\mathrm{BASE}}$-initialized language module via MLM using the text portion from the paired data. This optional step aims to adapt the language module to the speech domain to facilitate later alignment. After pre-training, the language module is discarded and only the speech module is used in downstream tasks.}
  \label{fig:architecture}
\end{figure*}

To solve these problems, we propose a novel SPeech-LAnguage joint pre-Training framework, SPLAT.
SPLAT contains a speech module and a language module for multi-modal understanding.
The speech module is a Transformer encoder trained from scratch and the language module is initialized from BERT.
Both modules leverage large-scale unannotated data for pre-training via masked language modeling.
In the speech module, each frame is seen as a token and is replaced with zero vector with a certain probability.
For each masked frame, we minimize the L1-distance between the predicted frame and the original frame.

Then, to make the speech module aware of the contextual information extracted from the language module, we design an alignment loss to align the representations from both modules in a shared latent semantic space.
In detail, we propose two alignment methods, a sequence-level one and a token-level one, that leverage a small amount of paired speech and text to minimize the disparity between the acoustic representations from the speech module and the textual representations from the language module.
In this way, the speech representations will carry not only the acoustic information but also the contextual knowledge from the text.
After this alignment, when text input is absent during fine-tuning, the speech module alone can produce representations that bridge the speech input and the language understanding output.

We conduct extensive evaluations on several downstream SLU tasks, including Fluent Speech Commands for intent detection, Switchboard for dialog act classification, CMU-MOSEI for spoken sentiment analysis, and Spoken SQuAD for spoken question answering.
SPLAT achieves superior results in all datasets.
For example, SPLAT improves the previous state-of-the-art performance on the Spoken SQuAD dataset by more than~10\%.
Furthermore, we show that SPLAT can perform well even given just a tiny portion of the labeled training data in downstream tasks.

\section{Related Work}
\label{sec:rw}
\paragraph{Spoken language understanding}
In recent years, due to its flexibility and effectiveness, end-to-end spoken language understanding~(SLU) has been proposed and applied to various tasks~\citep{qian2017exploring,serdyuk2018towards,lugosch2019speech}.
For instance, \citet{qian2017exploring} use an auto-encoder to initialize the SLU model.
\citet{lugosch2019speech} pre-train the model to recognize words and phonemes, and then fine-tune it on downstream tasks.
\citet{chen2018spoken} pre-train the model to categorize graphemes, and the logits are fed into the classifier.
In most of these approaches, the model pre-training requires annotated speech, e.g., word or phonemes corresponding to audio signals.
As a result, the massive unlabeled speech data cannot be utilized by these models.

\paragraph{Self-supervised pre-training for language}
Pre-trained models have achieved great success in both language and speech domains.
In language, BERT~\citep{devlin2019bert}, RoBERTa~\citep{liu2019roberta}, UniLM~\citep{dong2019unified}, and BART~\citep{lewis2020bart} have been successfully applied to natural language inference~\citep{zhang2020semantics}, question answering~\citep{zhu2018sdnet}, and summarization~\citep{zhu2019make}.
These pre-trained models leverage self-supervised tasks such as masked language modeling~(MLM), next sentence prediction, and de-noising autoencoder.

\paragraph{Self-supervised pre-training for speech}
In speech, wav2vec~\citep{schneider2019wav2vec} leverages contrastive learning to produce contextual representations for audio input; vq-wav2vec~\citep{baevski2020vq} and wav2vec~2.0~\citep{baevski2020wav2vec} further propose to discretize the original continuous audio signals in order to enable more efficient MLM training with Transformer~\citep{vaswani2017attention}.
Pre-trained speech models have been applied to ASR~\citep{ling2020deep,chung2020generative,baevski2020wav2vec}, phoneme recognition~\citep{song2020speech,liu2020non}, speech translation~\citep{nguyen2020investigating,chung2019towards}, and speech synthesis~\citep{chung2019semi}, to name a few.

Nevertheless, an SLU model must incorporate both acoustic and language understanding capabilities to project speech signals to semantic outputs.
Thus, a pre-trained model for SLU needs to address tasks beyond a single modality.

\paragraph{Speech and language joint pre-training}
Recently, SLU applications have prompted joint pre-training on both speech and text data.
SpeechBERT~\citep{chuang2020speechbert} applies MLM to pairs of audio and transcripts.
However, there are several crucial differences to compared to our work.
First, SpeechBERT contains a phonetic-semantic embedding module that requires forced alignment to first segment speech into word segments to obtain.
Second, both the pre-training and fine-tuning phases of SpeechBERT require both speech and text input, since it is designed for a specific spoken question answering task.
However, many SLU tasks only take speech as input, which does not align with the design of SpeechBERT.
In contrast, our model can learn to align acoustic and textual representations using just~(a small amount of) paired data during pre-training, and only needs speech input for downstream tasks.

\citet{denisov2020pretrained} propose to align speech and language embeddings in a method similar to ours.
However, there are several key differences.
First, \citet{denisov2020pretrained} employ the encoder of a pre-trained ASR model, which already requires plentiful of annotated speech to obtain.
Our model, on the other hand, conducts self-supervised learning to pre-train the speech module using unannotated speech.
Secondly, besides sequence-level alignment, we propose a token-level alignment method, which is suitable for token-level downstream tasks.
Last but not least, our model uses a much smaller paired speech and text for alignment~(10 hours) than~\citet{denisov2020pretrained}~(1,453 hours), yet still largely outperforms their method in intent detection and dialog act classification.

\section{Method}
\label{sec:method}
In this section we present SPLAT, a framework for learning joint contextual representations of speech and language.
The model consists of a speech module and a language module that share a similar architecture and learning algorithm.
The pre-training of SPLAT is divided into two steps.
First, we individually pre-train the speech and language modules using unannotated speech and text, respectively.
Then, we leverage a simple yet effective alignment task that uses only a small amount of paired speech and text data to align the representations from both modules in a shared latent semantic space such that the information learned by the language module is transferred to the speech module.
After pre-training, the language module is discarded and only the speech module is used in downstream tasks.

Below we formally describe the procedures for pre-training the speech~(§\ref{sec:speech_module}) and language modules~(§\ref{sec:language_module}), and the alignment loss~(§\ref{sec:align_module}) for aligning the representations from the two modules.
Figure~\ref{fig:architecture} provides an overview of the pre-training procedures of SPLAT.

\begin{figure*}[htbp]
  \centering
  \includegraphics[trim=5.6cm 7cm 0.2cm 9.5cm, clip,width=1.00\textwidth]{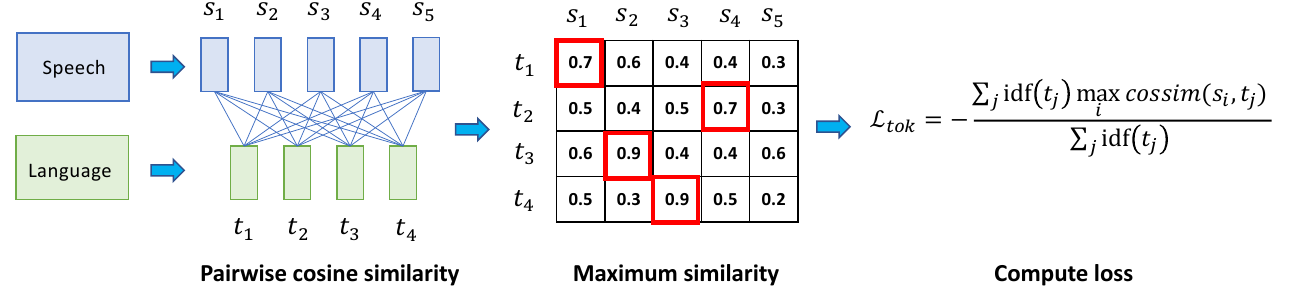}
  \caption{Token-level alignment between speech and language modules. $(\bs_{1}, ..., \bs_{5})$ are the output embeddings of the speech module and $(\bt_{1}, ..., \bt_{4})$ are those of the language module.}
  \label{fig:lseq}
\end{figure*}

\subsection{Speech module pre-training}
\label{sec:speech_module}
The goal of this module is to leverage unlabeled speech data to learn representations that capture meaningful acoustic information about speech utterances such as their phonetic content and speaker characteristics.
Formally, the input to the speech module is a~80-dimensional log Mel spectrogram, $(\bx_1, ..., \bx_n)$, where $\bx_{i}\in \mathbb{R}^{80}, 1\leq i \leq n$.
The speech module, which is implemented as a Transformer architecture, then produces hidden representations~$(\bs_1, ..., \bs_n)$ and predictions~$(\hat{\bx}_1, ..., \hat{\bx}_n)$, where~$\bs_{i}\in \mathbb{R}^{768}$ and~$\hat{\bx}_{i}\in \mathbb{R}^{80}$.

To boost its capacity for contextual understanding, we borrow the idea of masked language modeling~(MLM)~\citep{devlin2019bert,liu2020mockingjay,wang2020unsupervised,liu2020tera}.
Specifically, each audio frame~$\bx_i$ is replaced with a zero vector with a probability of~15\%.
The corresponding output~$\hat{\bx}_i$ is trained to be close to the original frame~$\bx_i$ via minimizing their L1-distance.
Additionally, since consecutive frames are highly correlated, it is possible that the model simply utilizes the local smoothness of speech signals for reconstructing a single frame and thus fails to capture useful information.
To avoid such issue, when a frame~$\bx_{i}$ is selected to be masked, its following three frames $\bx_{i + 1}$, $\bx_{i + 2}$, and $\bx_{i + 3}$ are also masked, and the model is asked to reconstruct all these masked frames.

Furthermore, according to SpecAugment~\citep{park2019specaugment}, the input features $(\bx_1, ..., \bx_n)$ can be seen as comprising two dimensions: time, i.e., the subscript~$i$, and channel, i.e., the elements in each~$\bx_i$.
While conventional MLM masks along certain time steps, the input signals can also be masked along the channel dimension.
In other words, each column vector $[\bx_{1,j}, ..., \bx_{n,j}]$ for $1\leq j \leq 80$ has a~15\% of chance to be masked, i.e., replaced with a zero vector.
This channel masking is combined with temporal masking to reinforce the model's capability to utilize contextual information from both time and channel, and reduce the impact of co-adaptation between acoustic frames.
The final pre-training objective for the speech module is to reconstruct the entire input sequence from the altered version of it: 
\begin{equation}
  \mathcal{L}_{sp}=\sum_{i = 1, 2, ..., n}\|\bx_i - \hat{\bx}_i\|_1
  \label{eq:lsp}
\end{equation}

We use the speech portion of the {train-clean-360} subset from the LibriSpeech corpus~\citep{panayotov2015librispeech} to pre-train the speech module, i.e., to minimize $\mathcal{L}_{sp}$.
This subset contains~360 hours of read speech produced by~921 speakers.
We follow the standard Kaldi setting, using a frame size of 25ms and a time shift of 10ms for generating the 80-dimensional log Mel spectrograms.
The spectrograms are normalized to zero mean and unit variance per speaker.

\subsection{Language module pre-training}
\label{sec:language_module}
The language module aims to offer contextual understanding for text input.
We directly employ the BERT$_{\mathrm{BASE}}$ model released by~\citet{devlin2019bert}, which is pre-trained on a large text corpus with the MLM task and contains rich textual representations, as the language module.
We denote the cross-entropy loss for the language MLM task as~$\mathcal{L}_{text}$.

Given input token embeddings $(\by_1, ..., \by_m)$, where $\by_{1}$ corresponds to the [CLS] token, the module produces contextual representations $(\bt_1, ..., \bt_m)$, where $\bt_{j}\in \mathbb{R}^{768}, 1\leq j \leq m$.

\subsection{Aligning speech and language representations}
\label{sec:align_module}
The input to most SLU tasks consists of only audio signals, but the model is required to conduct semantic understanding, which can be best handled when textual information is present.
Therefore, we propose to align the pre-trained speech and language representations in a shared semantic latent space.

\begin{table*}[htbp]
  \centering
  \caption{Variants of SPLAT. An \xmark\ indicates that the variant does not incorporate this step during pre-training. The step numbers correspond to those listed in Algorithm~\ref{alg:pretraining}.}
  \label{tab:model_ablations}
  \resizebox{\textwidth}{!}{
    \begin{tabular}{l|cccc}
      \toprule
      Model variant &  \makecell{Step 1. Pre-train\\speech module} &  \makecell{Step 2. Pre-train\\language module}  &  \makecell{Step 3. Adapt language\\module before alignment}  &  \makecell{Step 4. Type of\\alignment loss} \\
      \hline
      {\bf SPLAT-Scratch}    &  \xmark  &  \xmark  &  \xmark  &  \xmark \\
      {\bf SPLAT-Speech}     &  \cmark  &  \xmark  &  \xmark  &  \xmark \\
      {\bf SPLAT-Seq}        &  \cmark  &  \cmark  &  \xmark  &  $\mathcal{L}_{seq}$ \\
      {\bf SPLAT-Seq-MLM}    &  \cmark  &  \cmark  &  \cmark  &  $\mathcal{L}_{seq}$ \\
      {\bf SPLAT-Tok}        &  \cmark  &  \cmark  &  \xmark  &  $\mathcal{L}_{tok}$ \\
      {\bf SPLAT-Tok-MLM}    &  \cmark  &  \cmark  &  \cmark  &  $\mathcal{L}_{tok}$ \\
      \bottomrule
  \end{tabular}
  }
\end{table*}

Suppose a pair of speech and text data consisting of an acoustic feature sequence $(\bx_1, ..., \bx_n)$ and its transcript $(\by_1, ..., \by_m)$.
The speech and language modules separately produce the output representations $(\bs_1, ..., \bs_n)$ and $(\bt_1, ..., \bt_m)$.
We then propose two methods to align the embeddings from the modules: sequence-level and token-level alignment.

\paragraph{Sequence-level alignment}
For sequence-level alignment, we treat the first embeddings from the two output representations, i.e., $\bs_{1}$ and $\bt_{1}$, as the sequence-level representations of their respective sequences, and minimize their L1-distance:
\begin{equation}
  \mathcal{L}_{seq}=\|\bs_1 - \bt_1\|_1
  \label{eq:seq}
\end{equation}
Since our goal is to transfer the textual knowledge contained by the language module to the speech module, we only update the speech module to minimize $\mathcal{L}_{seq}$ and keep the language module fixed.

After pre-training, when the transcript is absent in downstream tasks, the first output embedding of the speech module $\bs_{1}$ will still be close to its corresponding text embedding $\bt_{1}$ from the language module, as if the transcript were given.
It follows that $\bs_{1}$ can then be used to predict the property of the whole audio input, e.g., intent classification.

\paragraph{Token-level alignment}
To achieve a finer level of alignment, each audio feature should be compared with its each text token.
Although forced alignment~\citep{gorman2011prosodylab} can establish this correspondence between audio signals and individual words, it requires a pre-trained ASR system to obtain.
Here we propose a method that automatically aligns audio features with textual tokens.

Inspired by BERTScore~\citep{zhang2020bertscore}, for each output text embedding~$\bt_j$, we first compute its cosine similarity with each output acoustic embedding~$\bs_i$, and select the acoustic feature with the highest similarity.
Then, the alignment is performed by maximizing the sum of these maximum similarities over all tokens, weighted by each token's inverse document frequency~(idf) to reduce the impact of common words:
\begin{equation}
  \mathcal{L}_{tok}=-\frac{\sum_{j=1}^m{\text{idf}(\bt_j)\max_i\text{cossim}(\bs_i, \bt_j)}}{\sum_{j=1}^m{\text{idf}(\bt_j)}}
  \label{eq:tok}
\end{equation}
The token-level alignment loss is illustrated in Figure~\ref{fig:lseq}.
Same as~$\mathcal{L}_{seq}$, when minimizing~$\mathcal{L}_{tok}$, the language module is kept fixed and only the speech module is updated.

To minimize the alignment loss, we randomly sample~10 hours of audio paired with its transcripts from the {train-clean-360} subset, of which the speech portion is used to pre-train the speech module~(§~\ref{sec:speech_module}).
In practice, before minimizing the alignment loss, we find it beneficial to train~(i.e., minimize~$\mathcal{L}_{text}$) the language module initialized with BERT$_{\mathrm{BASE}}$ with the~10-hour LibriSpeech transcripts with the MLM task.
This step allows the model to adapt to the speech domain and facilitates the following alignment task.

\begin{algorithm}[t]
  \renewcommand{\algorithmicrequire}{\textbf{Input:}}
  \renewcommand{\algorithmicensure}{\textbf{Output:}}
  \caption{Pre-training SPLAT}
  \label{alg:pretraining}
  \begin{algorithmic}[1]
  \REQUIRE An unlabeled speech corpus $\mathcal{X} = \{\bx^{(p)}\}^{N}_{p=1}$, an unlabeled text corpus $\mathcal{Y} = \{\by^{(q)}\}^{M}_{q=1}$, and a paired speech-text corpus $\mathcal{Z} = \{(\bx^{(k)}, \by^{(k)})\}^{K}_{k=1}$, where $K\ll N, M$.
  \STATE Use~$\mathcal{X}$ to train the speech module by minimizing~$\mathcal{L}_{sp}$~(Equation~\ref{eq:lsp}).
  \STATE Use~$\mathcal{Y}$ to train the language module by minimizing~$\mathcal{L}_{text}$~(we directly employ BERT$_\mathrm{BASE}$ from \citet{devlin2019bert} for this step).
  \STATE Use~$\{\by^{(k)}\}^{K}_{k=1}$ from~$\mathcal{Z}$ to train the language module by minimizing~$\mathcal{L}_{text}$.
  \STATE Use~$\mathcal{Z}$ to align the two modules by minimizing~$\mathcal{L}_{seq}$~(Equation~\ref{eq:seq}) or~$\mathcal{L}_{tok}$~(Equation~\ref{eq:tok}).
  \STATE Discard the language module.
  \ENSURE The final speech module.
  \end{algorithmic}
\end{algorithm}

\begin{table*}[htbp]
  \centering
  \caption{Summary of SLU datasets. For the rows of Train, Validation, and Test, the numbers indicate the number of utterances in the split.}
  \normalsize
  \label{tab:datasets}
  \begin{tabular}{l|cccc}
    \thickhline
    Task  &  \makecell{Intent\\detection}  &  \makecell{Dialog act\\classification}  &  \makecell{Spoken sentiment\\analysis}  &  \makecell{Spoken question\\answering} \\
    \hline
    Dataset          &  FSC  &  SwBD  &  CMU-MOSEI  &  Spoken SQuAD  \\
    \hline
    Num. of classes  &  31      &   42     &  7       &  -       \\
    Train/val/test   &  23.1k/3.1k/3.8k  &  97.8k/8.6k/2.5k &  16.2k/1.8k/4.6k &  35.1k/2.0k/5.4k \\
    \thickhline
  \end{tabular}
\end{table*}

We summarize the complete procedure of pre-training SPLAT in Algorithm~\ref{alg:pretraining}.
After pre-training, the language module is discarded and only the speech module is used in downstream tasks.

\section{Experiment Setup}
\label{sec:exp}
\subsection{Baselines}
We include a number of strong baselines from recent literature for each downstream task~\citep{lugosch2019speech,duran2018probabilistic,ghosal2018contextual,chuang2020speechbert}.
We also compare with another speech-language joint pre-training framework~\citep{denisov2020pretrained}.
For each baseline, the reported performance is achieved by system that either uses similar or more amounts of data than our model.

To verify the effectiveness of each component in SPLAT, we experiment with the following variants of it, including whether to pre-train the model, whether to use the language module and which alignment task to apply.
Table~\ref{tab:model_ablations} summarizes the considered model variants.
\begin{itemize}
  \item \textbf{SPLAT-Scratch}: No pre-training is conducted at all. Speech module is trained from scratch on downstream tasks.
  \item \textbf{SPLAT-Speech}: Only the speech module is pre-trained. Language module and alignment loss are not incorporated.
  \item \textbf{SPLAT-Seq}: SPLAT with sequence-level alignment loss $\mathcal{L}_{seq}$, but language module is not trained on LibriSpeech transcripts with MLM before alignment.
  \item \textbf{SPLAT-Seq-MLM}: SPLAT with sequence-level alignment loss $\mathcal{L}_{seq}$, and language module is trained on LibriSpeech transcripts with MLM before alignment.
  \item \textbf{SPLAT-Tok}: SPLAT with token-level alignment loss $\mathcal{L}_{tok}$, but language module is not trained on LibriSpeech transcripts with MLM before alignment.
  \item \textbf{SPLAT-Tok-MLM}: SPLAT with token-level alignment loss $\mathcal{L}_{tok}$, and language module is trained on LibriSpeech transcripts with MLM before alignment.
\end{itemize}

The speech module of SPLAT is a 3-layer Transformer encoder where each layer has a hidden size of~768 and~12 self-attention heads.
The language module is directly initialized from the pre-trained BERT$_{\mathrm{BASE}}$ released by~\citet{devlin2019bert}.


\subsection{Downstream SLU Tasks}
We evaluate our model on four different SLU applications: intent detection, dialog act classification, spoken sentiment analysis, and spoken question answering.
The first three belong to multi-class classification tasks, and the last one is a span prediction problem, which will be described in more detail below.
Table~\ref{tab:datasets} summarizes the used dataset for each application.
For all datasets, we use~80-dimensional log Mel spectrograms as input acoustic features as in the pre-training stage.

\paragraph{Intent detection}
We use the Fluent Speech Commands corpus~(FSC)~\citep{lugosch2019speech} for intent detection, where the goal is to correctly predict the intent of an input utterance.
In this dataset, each utterance is annotated with three slots: action, object, and location, where each slot can take one of multiple values.
The combination of slot values is defined as the intent of the utterance, and there are~31 unique intents in total.
In this work we follow the original paper to formulate intent detection as a simple~31-class classification task.

\begin{table*}[t]
  \centering
  \caption{Results on all downstream datasets. All numbers of our models are an average of three runs, of
which variances are negligibly small and not included. The metric is classification accuracy for FSC, SwBD and CMU-MOSEI. The metric for Spoken SQuAD is Audio Overlapping Score~(AOS).}
  \label{tab:full_set}
  \normalsize
  \begin{tabular}{l|cccc}
    \thickhline
    Model                    &  FSC   &  SwBD  &  CMU-MOSEI  &  Spoken SQuAD  \\
    \thickhline
    \rowcolor[gray]{0.95}
    Ours & & & & \\
    \thickhline
    {\bf SPLAT-Scratch}         &  97.6  &  65.8  &  68.8  &  30.4  \\
    {\bf SPLAT-Speech}          &  99.5  &  67.5  &  69.0  &  57.7  \\
    {\bf SPLAT-Seq}             &  99.5  &  74.6  &  72.5  &  62.7  \\
    {\bf SPLAT-Seq-MLM}         &  99.5  &  \textbf{76.3}  &  74.7  &  \textbf{65.9}  \\
    {\bf SPLAT-Tok}             &  99.2  &  71.2  &  70.4  &  58.0  \\
    {\bf SPLAT-Tok-MLM}         &  99.2  &  72.7  &  71.2  &  63.8  \\
    {\bf SPLAT-Seq-MLM 1-hour}  &  99.5  &  75.8  &  65.3  &  65.3  \\
    \thickhline
    \rowcolor[gray]{0.95}
    Baselines & & & & \\
    \thickhline
    \citet{lugosch2019speech}       &  98.8  &   -    &  -  &  -     \\
    \citet{duran2018probabilistic}  &    -   &  75.5  &  -  &  -     \\
    \citet{ghosal2018contextual}    &    -   &   -    &  \textbf{75.9}  &  -  \\
    \citet{chuang2020speechbert}    &    -   &   -    &  -  &  59.7  \\
    \citet{denisov2020pretrained}   & \textbf{95.5}  &  60.2  &  -  &  - \\
    \thickhline
  \end{tabular}
\end{table*}

\paragraph{Dialog act classification}
We use the NTX-format Switchboard corpus~(SwDA)~\citep{calhoun2010nxt}, a dialog corpus of~2-speaker conversations.
The goal is to correctly classify an input utterance into one of the~42 dialog acts.

\paragraph{Spoken sentiment analysis}
We use the CMU-MOSEI dataset~\citep{zadeh2018multimodal}, where each utterance is annotated for a sentiment score on a~$[-3, 3]$ Likert scale: [-3: highly negative, -2: negative, -1: weakly negative, 0: neutral, +1: weakly positive, +2: positive, +3: highly positive].
We treat the task as a~7-class classification problem.
And we only use audio signals in the input data.

For the above three tasks, during fine-tuning, an MLP network with one hidden layer of~512 units is appended on top of the speech module.
It converts the output representation of the first frame, i.e., $\bs_{1}$, for class prediction.
Both the pre-trained speech module and the randomly initialized MLP are fine-tuned on the training set for~10 epochs with a batch size of~64 and a fixed learning rate of~3e-4.
We compute classification accuracy after each training epoch and pick the best-performing checkpoint on the validation set to report results on the test set.

\paragraph{Spoken question answering}
We use the Spoken SQuAD dataset~\citep{li2018spoken}, which is augmented%
\footnote{\citet{li2018spoken} used Google text-to-speech to generate the spoken version of the articles in SQuAD.} from SQuAD~\citep{rajpurkar2016squad} for spoken question answering.
The model is given an article in the form of speech and a question in the form of text.
The goal is to predict a time span in the spoken article that answers the question.
In other words, the model outputs an audio segment extracted from spoken article as the answer.
The model is evaluated by Audio Overlapping Score~(AOS)~\citep{li2018spoken}: the greater the overlap between the predicted span and the ground-truth answer span, the higher the score will be.

During fine-tuning, given a spoken article and a question in the text form, the pre-trained speech module extracts audio representations of the article and pass them to a randomly initialized~3-layer Transformer encoder along with the tokenized textual question as input.
The Transformer then uses the self-attention mechanism to implicitly align elements of the input audio and textual features.
For each time step of the audio input, the Transformer is trained to predict whether this is the start of the span with a simple logistic regression.
A separate classifier is used for predicting the end of the span.

\section{Results and Analysis}
\label{sec:results}
\subsection{Main results}
Table~\ref{tab:full_set} shows the performance of models on all four downstream tasks.
Each number from our model is an average over three runs.
Based on the results, we make the following observations.

Firstly, compared with {\bf SPLAT-Scratch}, all pre-trained models achieve superior results, especially more than~30\% gain on Spoken SQuAD, proving the effectiveness of pre-training.

\begin{figure*}[t]
  \centering
  \includegraphics[width=1.0\textwidth]{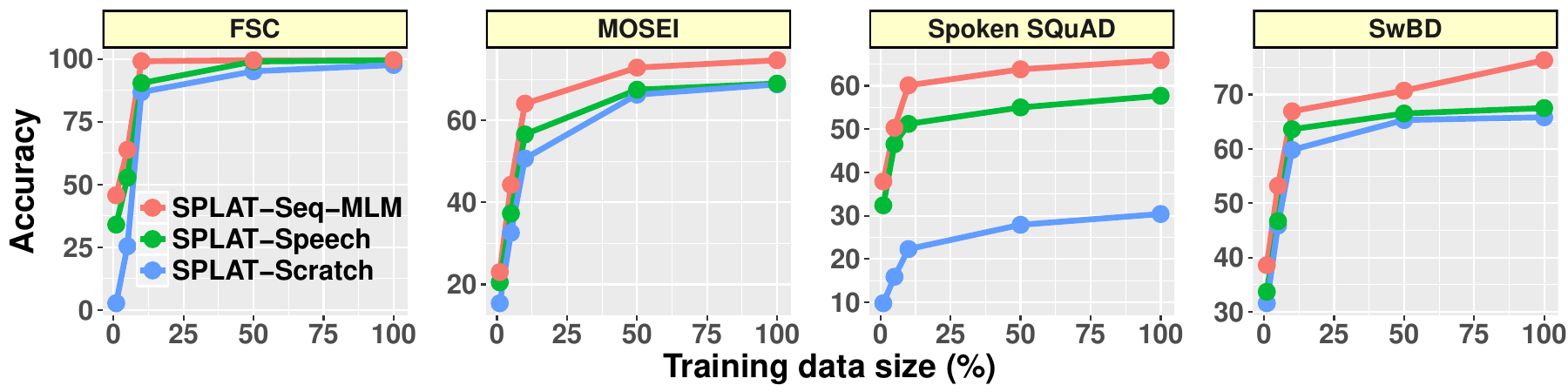}
  \caption{Performance on downstream tasks with varying training data sizes. All numbers are an average of three runs, of which variances are negligibly small and not included.}
  \label{fig:vary}
\end{figure*}

Secondly, the inclusion of language module and the alignment task during pre-training is very beneficial.
For instance, on CMU-MOSEI, {\bf SPLAT-Seq-MLM} outperforms {\bf SPLAT-Speech} by~5.7\%, and outperforms several baseline systems from recent literature.
We argue that as SLU tasks require the model to interpret acoustic signals and their underlying semantics, the language module will guide the speech module towards a mutual understanding of both modalities via our alignment task.

Thirdly, updating the language module using MLM during pre-training is helpful.
Although the language module has been initialized with BERT, adaptation to the speech domain can help with semantic understanding in the downstream task.

\paragraph{Types of alignment}
Comparing {\bf SPLAT-Seq} against {\bf SPLAT-Tok}, we find that sequence-level alignment outperforms token-level alignment on all four tasks, although the latter is supposed to learn more fine-grained multi-modal representations.
We leave the investigations of reasons for such phenomenon and more advanced token-level alignment approaches for future work.

\paragraph{Low-resource scenario}
We experiment with a version of SPLAT that uses only~1 hour of transcribed speech randomly sampled from the LibriSpeech {train-clean-360} subset for aligning speech and language modules, denoted as {\bf SPLAT-Seq-MLM 1-hour}.
The language module of {\bf SPLAT-Seq-MLM 1-hour}---after being initialized with BERT$_\mathrm{BASE}$---is trained on the~1-hour LibriSpeech transcripts before minimizing the alignment loss.
It achieves comparable results with the best variant {\bf SPLAT-Seq-MLM}: same accuracy on FSC, 0.5\% less on SwBD, and 0.6\% less on Spoken SQuAD.
This shows that with a small amount of labeled speech data, our pre-training framework can achieve good results on downstream tasks.

\subsection{Robustness to the size of downstream training data}
As human labeling is time-consuming and labor-intensive, the amount of labeled training data for downstream tasks is often small and insufficient.
In this section, we show that with effective pre-training, the model will be less dependent on the amount of downstream labeled data.

We randomly sample 50\%, 10\%, 5\%, and 1\% of the training data in the downstream tasks, and evaluate the performance of different variants of SPLAT when fine-tuned on the sampled data.

Figure~\ref{fig:vary} shows the performance on all four downstream tasks with varying training data sizes.
We observe that among the variants, {\bf SPLAT-Seq-MLM} is least sensitive to training data sizes.
For instance, in FSC, with only~10\% of the training data, its accuracy only drops~0.4 points.
In comparison, both {\bf SPLAT-Scratch} and {\bf SPLAT-Speech} drops about~10 points.
And the gaps are in general larger when the size of training data further shrinks.
Therefore, our proposed joint pre-training of speech and language modules can help the model quickly adapt to downstream tasks given a modest amount of training data.

\subsection{The geometry of the speech latent space before and after alignment}
So far we have empirically demonstrated the effectiveness of SPLAT for learning multi-modal speech-language representations that are useful in various SLU tasks.
Here we further show that our sequence-level alignment loss~(Equation~\ref{eq:seq}) can help project two speech utterances that have similar textual embeddings to nearby points in the speech latent space.
\begin{table}[htbp]
  \centering
  \caption{Average cosine similarity between all pairs of speech embeddings~($S_{avg}$), and the average cosine similarity between a speech embedding~$\bs_1^{(p)}$ and that of an utterance whose textual embedding is closest to the corresponding textual embedding~$\bt_1^{(p)}$~($S_{closest}$).}
  \label{tab:cos}
  \begin{tabular}{l|cc}
    \thickhline
    Model                &  $S_{avg}$  &  $S_{closest}$ \\
    \hline
    {\bf SPLAT-Speech}   &    0.136    &    0.238 \\
    {\bf SPLAT-Seq}      &    0.144    &    0.781 \\
    {\bf SPLAT-Seq-MLM}  &    0.148    &    0.829 \\
    \thickhline
  \end{tabular}
\end{table}

Recall that we use the embedding of the first token/feature to represent an utterance and conduct sequence-level alignment (Equation~\ref{eq:seq}).
Suppose~$\bt_{1}^{(p)}$ and~$\bs_{1}^{(p)}$ correspond to the textual and speech embeddings of the first utterance by SPLAT and~$\bt_{1}^{(q)}$ and~$\bs_{1}^{(q)}$ correspond to the embeddings of the second utterance.
Then, if~$\bt_{1}^{(p)}\approx\bt_{1}^{(q)}$, our SPLAT model trained with the sequence-level alignment loss will produce~$\bs_{1}^{(p)}\approx\bs_{1}^{(q)}$.

We use the dev-clean subset from the LibriSpeech corpus for the analysis.
First, we compute the average pairwise cosine similarity between the utterances of all speech embeddings:
\begin{equation}
    S_{avg} = \frac{1}{K(K-1)/2}\sum^{K}_{p=2}\sum^{p-1}_{q=1} \text{cossim}(\bs_1^{(p)}, \bs_1^{(q)}),
\end{equation}
where~$K$ is the number of utterances in dev-clean.

Next, for each utterance with its speech and textual embeddings denoted as~$\bs_1^{(p)}$ and~$\bt_1^{(p)}$ respectively, we first use~$\bt_1^{(p)}$ to retrieve the utterance with the most similar textual embedding~$\bt_1^{(q^{*})}$, i.e.,~$q^{*} = \text{argmax}_{1\leq q\leq K,q\neq p}\text{cossim}(\bt_1^{(p)}, \bt_1^{(q)})$.
We then compute the cosine similarity between~$\bs_1^{(p)}$ and~$\bs_1^{(q^{*})}$ and take the average of such value over all utterances in dev-clean:
\begin{equation}
    S_{closest} = \frac{1}{K}\sum^{K}_{p=1}\text{cossim}(\bs_1^{(p)}, \bs_1^{(q^{*})}).
\end{equation}
We show the~$S_{avg}$ and~$S_{closest}$ of embeddings produced by {\bf SPLAT-Speech}, {\bf SPLAT-Seq}, and {\bf SPLAT-Seq-MLM} in Table~\ref{tab:cos}.

We see that~$S_{avg}$ is approximately the same for all model variants.
However, $S_{closest}$, the average similarity between the speech embeddings of two linguistically similar utterances, increases from~0.238 to~0.781 after aligning the speech and language modules, and further increases to~0.829 after adapting the language module on LibriSpeech transcripts with MLM before the alignment.
Overall, SPLAT can make a pair of semantically similar utterances to have much closer speech embeddings, compared with other random pairs of utterances.

These results demonstrate that via an cross-modal alignment loss as simple as Equation~\ref{eq:seq}, SPLAT can effectively transfer knowledge from the language module to the speech module to capture both acoustic and linguistic information of speech utterances.

\section{Conclusions}
\label{sec:conclusion}
Spoken language understanding~(SLU) tasks require an understanding of the input audio signal and its underlying semantics.
In this paper, we present a novel speech-language joint pre-training framework, SPLAT, to carry out both speech and language understanding tasks during pre-training.
Besides a self-supervised training on the speech and language modules, we propose two methods to align the semantic representations from both modules using a modest amount of labeled speech data.
The speech module can quickly adapt to downstream tasks and achieve superior results on various SLU datasets including intent detection, dialog act classification, spoken sentiment analysis, and spoken question answering.
This joint pre-training also makes the model less sensitive to the amount of labeled training data in downstream domains.

For future work, we plan to integrate automatic speech recognition and natural language generation into our framework to achieve good results on spoken language generation tasks.

\bibliography{naacl2021}
\bibliographystyle{acl_natbib}



\end{document}